\documentclass[11pt]{article}
\usepackage{UF_FRED_paper_style}
\DeclareUnicodeCharacter{2009}{\,}

\title{Exploring Human Behavior During Abstract Rule Inference and Problem Solving with the Cognitive Abstraction and Reasoning Corpus}

\author[1,4]{Caroline Ahn\thanks{Corresponding author: ahncj@bu.edu}}
\author[1,3]{Quan Do}
\author[2,4]{Leah Bakst}
\author[2,4]{Michael P. Pascale}
\author[1,2,3,4]{Joseph T. McGuire}
\author[1,2,3]{Michael E. Hasselmo}
\author[1,2,3,4]{Chantal E. Stern}

\affil[1]{Graduate Program for Neuroscience, Boston University, Boston, MA, USA}
\affil[2]{Department of Psychological and Brain Sciences, Boston University, Boston, MA, USA}
\affil[3]{Center for Systems Neuroscience, Boston University, Boston, MA, USA}
\affil[4]{Cognitive Neuroimaging Center, Boston University, Boston, MA, USA}

\date{}
\begin{document}
{\setstretch{.8}
\maketitle
\begin{abstract}

Humans exhibit remarkable flexibility in abstract reasoning, and can rapidly learn and apply rules from sparse examples. To investigate the cognitive strategies underlying this ability, we introduce the Cognitive Abstraction and Reasoning Corpus (CogARC), a diverse human-adapted subset of the Abstraction and Reasoning Corpus (ARC) which was originally developed to benchmark abstract reasoning in artificial intelligence. Across two experiments, CogARC was administered to a total of 260 human participants who freely generated solutions to 75 abstract visual reasoning problems. Success required inferring input-output rules from a small number of examples to transform the test input into one correct test output. Participants’ behavior was recorded at high temporal resolution, including example viewing, edit sequences, and multi-attempt submissions. Participants were generally successful (mean accuracy $\approx90\%$ for experiment 1 ($n=40$), $\approx80\%$ for experiment 2 ($n=220$) across problems), but performance varied widely across problems and participants. Harder problems elicited longer deliberation times and greater divergence in solution strategies. Over the course of the task, participants initiated responses more quickly but showed a slight decline in accuracy, suggesting increased familiarity with the task structure rather than improved rule-learning ability. Importantly, even incorrect solutions were often highly convergent, even when the problem-solving trajectories differed in length and smoothness. Some trajectories progressed directly and efficiently toward a stable outcome, whereas others involved extended exploration or partial restarts before converging. Together, these findings highlight CogARC as a rich behavioral environment for studying human abstract reasoning, providing insight into how people generalize, misgeneralize, and adapt their strategies under uncertainty.

\noindent

\end{abstract}
}


\section{Introduction}
Humans excel at identifying abstract patterns and structures that can be flexibly applied to novel situations, supporting adaptive generalization across contexts. This predisposition is believed to be supported by inductive biases, which allow rules to be inferred from limited information. Such ability for inductive reasoning has long been studied in the field of psychology, with diverse models and theories such as Bayesian models, relevance theory, and connectionist models proposed to capture the mechanisms behind it \cite{tenenbaum_theory-based_2006, sloman_feature-based_1993, medin_relevance_2003, schum_evidential_1994, collins_reasoning_2012, do_neural_2021}. Despite this, significant questions remain about how humans learn to apply abstract rules, especially in ambiguous settings where multiple solutions may appear plausible. While inductive reasoning allows for fast learning in constrained environments, the uncertainty involved makes it susceptible to errors that arise from biases, heuristics, and misrepresentative data \cite{sloman_problem_2005, tversky_judgment_1974, hayes_who_2023}.

Traditionally, abstract reasoning has been studied using tasks such as Raven’s Progressive Matrices (RPM). In a typical RPM task, test-takers are given a 3x3 matrix of image panels which feature basic shapes that are organized according to an underlying rule, and are asked to pick the correct image for the last panel out of multiple choices. RPM is widely accepted as a robust measure of fluid intelligence in both clinical and research settings \cite{carpenter_what_2018, cipolotti_graph_2022, preusse_fluid_2011, mazhirina_ravens_2016}, and the relatively simple and adaptable task structure is suited for studying the brain activity underlying abstract reasoning using neuroimaging methods such as fMRI \cite{mazhirina_ravens_2016, prabhakaran_neural_1997, morin_functional_2022}. It has also been adapted for study across various disorders and life stages \cite{simard_autistic_2015, rommel_longitudinal_2015, smirni_ravens_2020, murphy_lifespan_2023}, and for assessing reasoning in AI \cite{nelson_minimal_2024, malkinski_deep_2025}.

However, while RPM remains a foundational paradigm in experimental psychology, its multiple-choice format and binary accuracy scoring offer a limited window into the cognitive processes involved in abstract rule learning, updating, and application. Such task structure may not fully capture the strategic aspects of problem-solving. Moreover, when it comes to testing AI, programs can often excel at RPM-style tasks through extensive dataset training, but this success frequently reflects overfitting to superficial statistical regularities and/or dataset-specific pattern recognition\cite{mitchell_abstraction_2021} rather than true abstract reasoning \cite{rohlfs_generalization_2022, barrett_measuring_2018, zhang_raven_2019}. To address these gaps, the Abstraction and Reasoning Corpus (ARC) was developed as a more open-ended benchmark of abstract visual reasoning in machines \cite{chollet_measure_2019}. In ARC, solvers are given a small set of input-output examples and must infer the transformation rules to generate a correct output for a novel test input (Fig. 1). The rules span a range of visual concepts grounded in core knowledge theory from developmental psychology \cite{spelke_core_2007}, and the task structure allows for more nuanced investigation of few-shot learning, real-time strategy, and error analysis. The full ARC dataset comprises 400 training problems and 400 evaluation problems, and each problem contains a novel rule.

Several recent studies have adapted ARC for use in human behavioral research, demonstrating its potential as a tool for studying cognition. For example, Johnson et al. (2021) asked participants to solve 10 ARC problems and provide natural language descriptions of the rules they inferred, finding overall strong performance with wide variation across tasks \cite{johnson_fast_2021}. LeGris et al. (2024, 2025) expanded this approach to a much larger sample, presenting 5 problems each from the ARC training or evaluation set and comparing response accuracy and confidence \cite{legris_comprehensive_2025, legris_h-arc_2024}. Acquaviva et al. (2021) introduced a communication-based version, LARC, where participants played the roles of describers and builders, testing the feasibility of solving ARC tasks via verbal instruction alone \cite{acquaviva_communicating_2021}, and ConceptARC (Moskvichev et al., 2023) curated a structured subset of problems organized by transformation type, enabling comparison across conceptual categories \cite{moskvichev_conceptarc_2023}.

Together, these studies highlight the richness of ARC as a behavioral tool and demonstrate the wide range of approaches that can be used to study human reasoning with this framework. Many have focused on benchmarking human performance against AI systems or collecting structured data for model development — important contributions that have helped establish ARC as a useful comparative tool. However, because participants in these studies typically solved only a handful of problems each, they offer limited insight into how reasoning strategies evolve within individuals over time, how rule inference adapts to different kinds of problems, or how participants converge on shared (even if incorrect) solutions. In short, the temporal dynamics of rule learning and problem solving remain largely unexplored.

To address these gaps, we developed the Cognitive Abstraction and Reasoning Corpus (CogARC) — a human-adapted version of ARC designed specifically to study the cognitive mechanisms of rule inference in greater depth. Building on the structure of the original problem set, we selected 75 diverse problems from the ARC training set to test across a wide range of rule complexity and conceptual variety. Each participant attempted the full set (or as many as possible within a two-hour session), allowing for rich within-participant comparisons over time and examination of across-participant heterogeneity within-problem. We captured detailed behavioral traces to reconstruct participants’ reasoning trajectories rather than relying solely on accuracy outcomes to measure performance. In addition to our empirical analyses reported here, the CogARC dataset is released as a public resource for studying human abstraction and generalization, learning from sparse examples, and error structure at scale.

Across two experiments — a supervised pilot study (Experiment 1, n=40) and a larger-scale online replication (Experiment 2, n=220) — we investigated how people learn and apply abstract rules over time. While participants were generally successful, performance varied widely across problems and was highly related to rule complexity. Harder problems elicited longer deliberation times, greater divergence in action trajectories, and more heterogeneity in submitted solutions, suggesting systematic challenges in rule inference. Over the course of the task, participants became faster at initiating solutions, but their accuracy showed a small but consistent decline, perhaps suggestive of a speed-accuracy tradeoff. Even when incorrect, participants often converged on similar solution paths and final outputs, revealing consistent structured misinterpretations.

\section{Methods}

\subsection{Experiment 1}
\subsubsection{Human participants}
Human-participant procedures were approved by an Institutional Review Board; informed consent was obtained from all participants. 40 participants were recruited via a University student job board. Eligible participants were between 18-35 years of age, with no history or current condition of neurological or psychiatric problems, not taking any psychoactive drugs, and with normal or corrected-to-normal vision. We did not collect demographic information for this sample. Participants received monetary compensation of \$30 for participation in the study. 

\subsubsection{Task design}
The task was presented to participants on a browser-based interface (Figure 1) with the experimenter supervising study participation through an encrypted Zoom call. Participants learned the input-output transformation rules by studying the examples on the left of the screen, then applied the rules to the test input (shown in the middle of the screen) by drawing out the test output in the grid editor to the right. They could select from ten colors and click and/or drag in the editor to fill out the tiles. They also had the option to copy the test input image to the test output grid, resize the grid, and to reset the grid to all black if they wanted to start over. We recorded response time (on the scale of milliseconds) on any action participants made during a trial. Participants solved the full set of 75 problems, presented in the same order for everyone. Participants were allowed up to three attempts on each trial with feedback. They were given two five-minute breaks, one after solving 25 problems and another after solving 50. A task progress indicator tracked participants’ progress through the problem set.

\begin{figure}[H]
    \centering
        \includegraphics[scale=0.9]{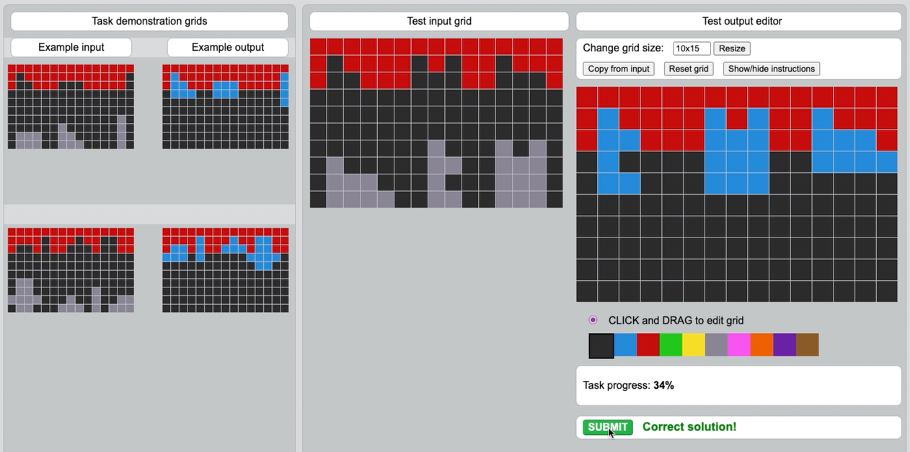}
    \caption{Task interface for experiment 1. The study participants solved 75 visuospatial abstract reasoning problems on the web browser-based interface shown above. The screen was divided into three parts: input-output examples to the left, test input grid in the middle, and a test output editor that participants could manipulate to the right.}
    \label{fig:1}
\end{figure}

Seventy-five problems were selected for human testing from the training set of the Abstraction and Reasoning Corpus (ARC) \cite{chollet_measure_2019}. Problems were manually reviewed by experimenters to ensure representation of a range of abstract rule types and difficulty levels, and to minimize overlap in rules so that each problem was novel to the participant. We only selected problems where the grid sizes were the same between input and output (Table 1). Three independent raters categorized each problem into one of four main core knowledge categories and rated the level of complexity (1-3) for each (Table 2). Complexity was determined by how many core knowledge concepts were used in the rule and the level of interaction between different elements of the rule (eg. conditional: color tiles in green \textit{if} there are fewer than three connected tiles in the same color, else keep as is). Disagreements were resolved by discussion. Complexity ratings were validated against participant performance on the task. See Figure 2 for an example of two different problems from the CogARC dataset with their respective task properties.

\begin{table}[H]
  \centering
  \caption{Range of task properties.}
  \label{tab:1}
  \scalebox{1.0}{    \begin{tabular}{rlr}
    \toprule
    \multicolumn{1}{c}{\textbf{Problem features}} & \multicolumn{1}{c}{\textbf{Minimum}} & \multicolumn{1}{c}{\textbf{Maximum}}      \\ \hline \hline
    \multicolumn{1}{l}{Grid size (tiles)} & \multicolumn{1}{c}{3x3} & \multicolumn{1}{c}{21x21}\\
    \hline
    \multicolumn{1}{l}{Number of input-output examples} & \multicolumn{1}{c}{2} & \multicolumn{1}{c}{6}  \\
    \hline
    \multicolumn{1}{l}{Number of colors used (excluding black)} & \multicolumn{1}{c}{1}& \multicolumn{1}{c}{9}  \\
    \bottomrule
    \end{tabular}}
\end{table}%

\begin{table}[H]
    \centering
    \caption{Problems by rule type.}
    \label{tab:2}    
    \scalebox{1.0}{    \begin{tabular}{rlr}
    \toprule
    \multicolumn{1}{c}{\textbf{Types of rules}} & \multicolumn{1}{c}{\textbf{Number of problems}}
     \\ \hline \hline
     \multicolumn{1}{l}{\textbf{Core knowledge categories}}& \multicolumn{1}{c}{}\\
     \multicolumn{1}{l}{Objectness} & \multicolumn{1}{c}{20}\\
     \multicolumn{1}{l}{Geometry and pattern} & \multicolumn{1}{c}{31}\\
     \multicolumn{1}{l}{Number and counting} & \multicolumn{1}{c}{15}\\
     \multicolumn{1}{l}{Goal-directedness} & \multicolumn{1}{c}{9}\\
     \hline
     \multicolumn{1}{l}{\textbf{Level of complexity}} & \multicolumn{1}{c}{}\\
     \multicolumn{1}{l}{1} & \multicolumn{1}{c}{21}\\
     \multicolumn{1}{l}{2} & \multicolumn{1}{c}{38}\\
     \multicolumn{1}{l}{3} & \multicolumn{1}{c}{16}\\
     \bottomrule
    \end{tabular}}
\end{table}%

\subsection{Experiment 2}
\subsubsection{Human participants}
Human-participant procedures were approved by an Institutional Review Board; informed consent was obtained from all participants. 233 participants were recruited via Amazon Mechanical Turk. Eligible participants were between 18-35 years of age, with no history or current condition of neurological or psychiatric problems, not taking any psychoactive drugs, and with normal or corrected-to-normal vision. Eligibility was assessed via a self-reported screening questionnaire. Participants also completed a demographics questionnaire (Table 2). Participants were not allowed to participate in the experiment more than once, so each problem was novel to them. After consenting to participate, participants were redirected to the study landing page. After completing 75 trials or participating in the study for 2 hours, participants were provided a unique code which they could then enter to the survey on Mechanical Turk to confirm task completion and receive payment. The base compensation for study participation was \$5, and participants were given a performance bonus in the range of \$0.10 to \$0.20 per problem solved correctly to encourage task engagement. Only 220 participants who were able to correctly solve more than ten problems were included in analysis for data quality control. We also excluded from data analysis problems that participants solved without viewing the input-output examples (see below), as that indicated they did not properly learn the rules.

\begin{table}[H]
  \centering
  \caption{Basic demographics.}
  \label{tab:3}
  \scalebox{0.8}{    \begin{tabular}{rlrrrrr}
    \toprule
    \multicolumn{1}{c}{\textbf{Demographics}} & \multicolumn{1}{c}{\textbf{Categories}} & \multicolumn{1}{c}{\textbf{n}} & \multicolumn{1}{c}{\textbf{Percentage (\%)}} & \multicolumn{1}{c}{\textbf{Mean}} & \multicolumn{1}{c}{\textbf{Median}}  & \multicolumn{1}{c}{\textbf{SD}} \\ \hline \hline
    \multicolumn{1}{l}{Age} & & 220& & 29.55& 31&4.06\\
    \hline
    \multicolumn{1}{l}{Gender} & Male& \multicolumn{1}{c}{115} & \multicolumn{1}{c}{52.3} & & & \\
          & Female& \multicolumn{1}{c}{104} & \multicolumn{1}{c}{47.3} & & & \\
          & Other& \multicolumn{1}{c}{1} & \multicolumn{1}{c}{0.5} & & & \\
    \hline
    \multicolumn{1}{l}{Native language} & English& \multicolumn{1}{c}{215} & \multicolumn{1}{c}{97.7} & & & \\
          & Other& \multicolumn{1}{c}{4} & \multicolumn{1}{c}{1.8} & & & \\
          & Not reported& \multicolumn{1}{c}{1} & \multicolumn{1}{c}{0.5} & & & \\
    \hline
    \multicolumn{1}{l}{Race/Ethnicity} & White& \multicolumn{1}{c}{150} & \multicolumn{1}{c}{68.2} & & & \\
          & Black& \multicolumn{1}{c}{37} & \multicolumn{1}{c}{16.8} & & & \\
          & Asian& \multicolumn{1}{c}{18} & \multicolumn{1}{c}{8.2} & & & \\
          & Other& \multicolumn{1}{c}{11} & \multicolumn{1}{c}{5.0} & & & \\
          & Not reported& \multicolumn{1}{c}{4} & \multicolumn{1}{c}{1.8} & & &\\
    \hline
    \multicolumn{1}{l}{Years of education} & & & & 15.3& 16&2.22\\
    \hline
    \multicolumn{1}{l}{Highest degree} & Bachelor's degree& \multicolumn{1}{c}{94} & \multicolumn{1}{c}{42.7} & & & \\
          & High school diploma or equivalency& \multicolumn{1}{c}{60} & \multicolumn{1}{c}{27.3} & & & \\
          & Master's degree& \multicolumn{1}{c}{35} & \multicolumn{1}{c}{15.9} & & &\\
          & Associate degree& \multicolumn{1}{c}{24} & \multicolumn{1}{c}{10.9}& & &\\
          & Doctorate& \multicolumn{1}{c}{3} & \multicolumn{1}{c}{1.4} & & &\\
          & Professional& \multicolumn{1}{c}{2} & \multicolumn{1}{c}{0.9} & & &\\
          & None of the above& \multicolumn{1}{c}{2} & \multicolumn{1}{c}{0.9} & & &\\
    \hline
    \multicolumn{1}{l}{Job status} & Working full-time& \multicolumn{1}{c}{160} & \multicolumn{1}{c}{72.7} & & &\\
        & Working part-time& \multicolumn{1}{c}{30} & \multicolumn{1}{c}{13.6} & & &\\
        & Keeping house full-time& \multicolumn{1}{c}{9} & \multicolumn{1}{c}{4.1} & & & \\
        & Unemployed& \multicolumn{1}{c}{8} & \multicolumn{1}{c}{3.6} & & &\\
        & Looking for work& \multicolumn{1}{c}{6} & \multicolumn{1}{c}{2.7} & & &\\
        & Full-time student& \multicolumn{1}{c}{6} & \multicolumn{1}{c}{2.7} & & &\\
        & Other& \multicolumn{1}{c}{1} & \multicolumn{1}{c}{0.5} & & &\\
    \bottomrule
    \end{tabular}}
\end{table}%

\subsubsection{Task design}
Participants solved the same 75 problems as in Experiment 1. The task was presented to participants on a browser-based interface that was comprised of an example view screen, where they learned the rules from the input-output examples, and an edit view screen, where they could see the test input and draw their solutions onto the test output grid using any of the ten colors in the color bar (Figure 3A). They also had the option to copy the test input image to the test output grid, and to reset the grid to all black if they wanted to start over. At the beginning of each problem, participants were shown the edit view screen by default, but they could freely switch back and forth between the example and edit views by clicking on a button.

All participant actions were logged with millisecond-level timestamps, allowing precise measurement of response times and edit sequences throughout each trial. Problem order was randomized across participants. Each problem allowed up to three attempts, with feedback provided after each attempt. Participants were shown their cumulative bonus earnings during the task. Participants were permitted to take self-timed breaks as needed. A progress indicator displayed their advancement through the problem set.

\begin{figure}[H]
    \centering
        \includegraphics[scale=0.7]{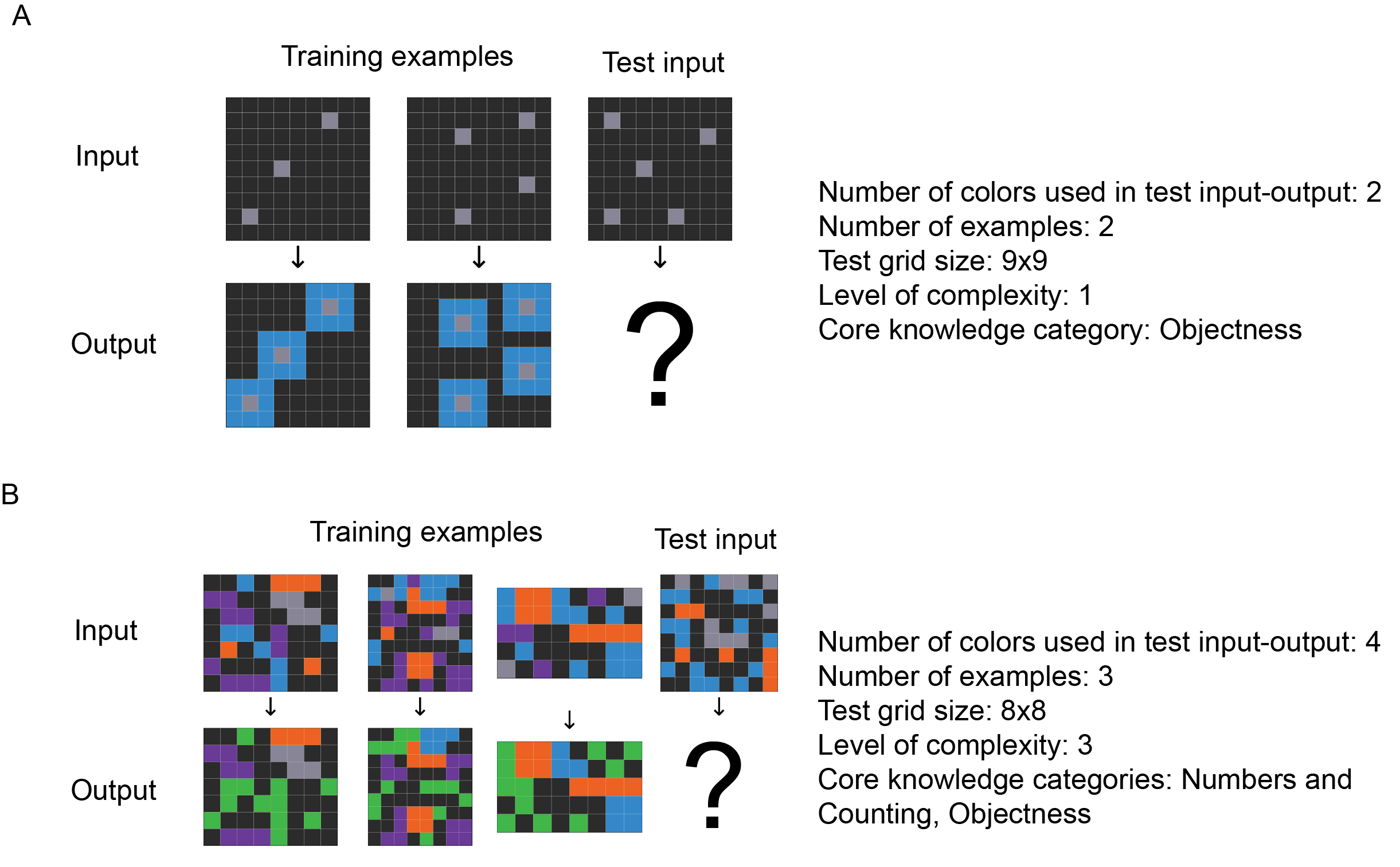}
    \caption{Two problems from the CogARC task and their task properties. Problem A is solved by drawing a blue outline around the grey tiles in the input. The core knowledge prior required to learn this rule is ‘objectness’. Problem B has a conditional rule in which any instances of fewer than 3 connected tiles of the same color in the input get re-colored to green in the output. Therefore, we can say this problem involves the core knowledge priors of ‘numbers and counting’ and ‘objectness’, and is of higher complexity than Problem A. For visualization purposes, problems are shown here in a simplified schematic format rather than the full interactive task interface used in the experiment.}
    \label{fig:2}
\end{figure}

\begin{figure}[H]
    \centering
        \includegraphics[scale=1.0]{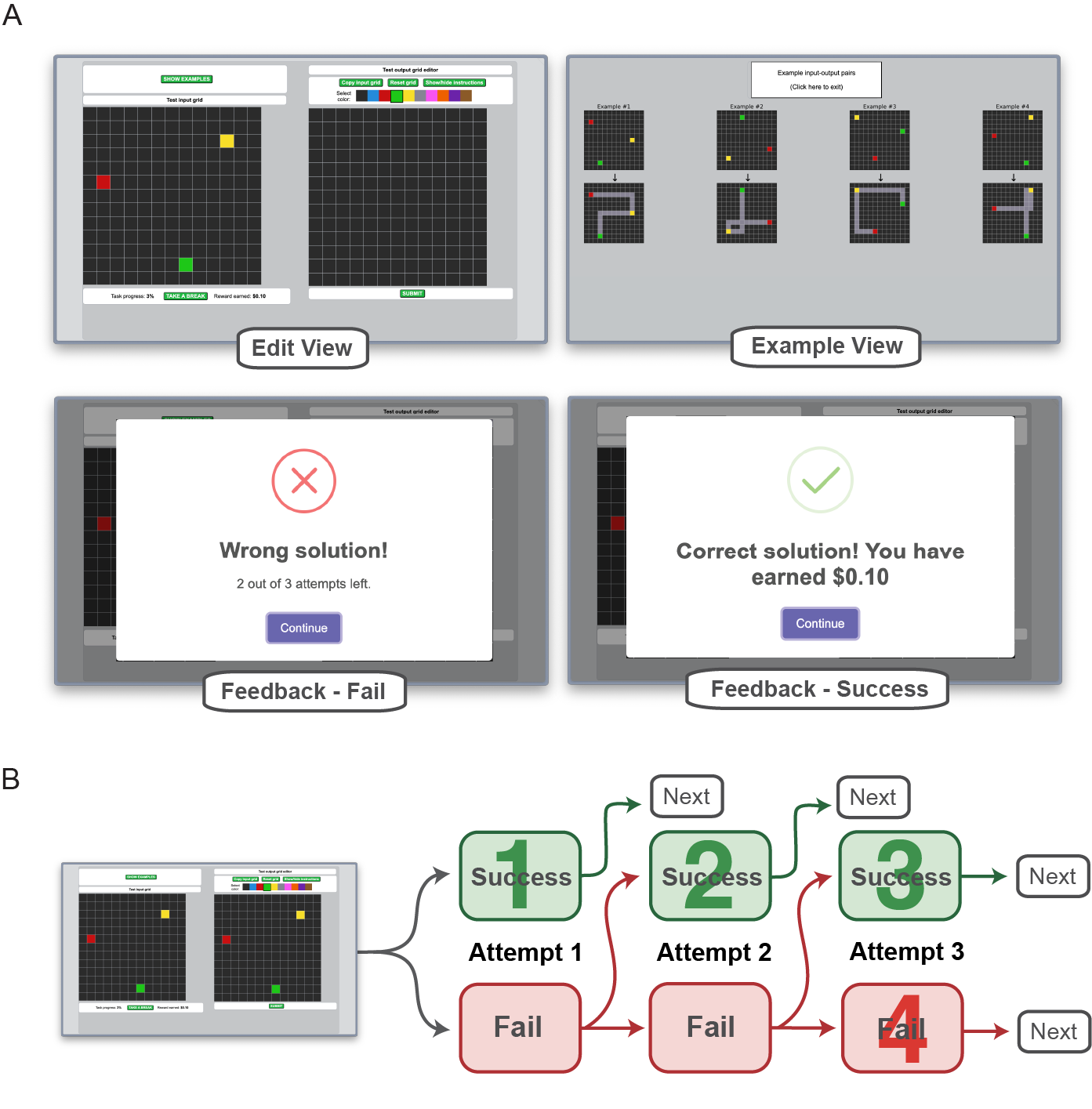}
    \caption{Task schematic. A) Experiment 2 task interface Participants could freely switch back and forth between edit view and example view. They were given feedback upon submission. B) Problem solving flow and difficulty scoring Participants got up to three attempts to reach the correct solution for a problem. Each participant was given a difficulty score of 1-4 for each problem based on how many submissions it took and what the outcome was.}
    \label{fig:3}
\end{figure}

\subsection{Behavioral analysis}
All analyses were performed using Python (version 3.10.9), with statistical calculations conducted using the $scipy.stats$ library. Data visualization was carried out using $seaborn$ and $matplotlib$.

\subsubsection{Performance measures}
We calculated participant accuracy on the overall task set by how many problems they solved correctly (across all three attempts) over the total number of problems they completed. We also measured participant performance on each problem by calculating a difficulty score based on the number of attempts made and whether the final outcome was a success. The difficulty scores ranged from 1 (success on first attempt) to 4 (all attempts failed) (Figure 3B).

We also looked at the amount of time participants spent between seeing a new problem and making their first edit action on the output grid (deliberation time).

\subsubsection{Edit Sequence Analysis}
To investigate the fine-grained dynamics of participant problem-solving strategies, we analyzed the edit sequences observed during each trial. An edit sequence refers to the collection of actions a participant took to modify the output grid while attempting to solve a given problem. Each edit is defined by a tuple containing:
\begin{itemize}
    \item the x and y coordinates of the modified tile,
    \item the color applied during that edit.
\end{itemize}
To measure how similarly participants approached each problem, we computed the pairwise Jaccard similarity between all participant edit sequences within each problem. The Jaccard similarity compares the degree of overlap between two sets—here, the sets of (x, y, color) edits made by each participant pair—by dividing the size of the intersection by the size of the union \cite{real_probabilistic_1996}:
\begin{equation}
Jaccard(A,B) = \frac{|A \cap B|}{|A \cup B|}
\end{equation}
Because this metric operates on unordered sets, it is agnostic to the temporal ordering of edits and naturally accounts for differences in sequence length. Even if one participant made more edits than another, the similarity score reflects the proportional overlap in their editing decisions rather than being biased by raw sequence length. By averaging the pairwise Jaccard similarities across all participant pairs for a given problem, we obtained a measure of how convergent or divergent participant solution strategies were at the level of inidivudal grid edits.

\subsubsection{Trends over time}
All participants solved the problems in a fully randomized order. To determine if there were any effects of learning or training over the course of the task, we analyzed trends in performance measures over time. We performed a linear regression analysis on each participant’s data, using serial trial number as the independent variable and example learning time as the dependent variable. This resulted in an individual regression coefficient for each participant, representing the rate of change in example learning time across trials. We performed the same analysis on difficulty score as a function of serial trial number. We then tested the coefficients against zero at the group level using one-sample t-tests.

\subsubsection{Shared errors}
We identified up to three most common errors for first attempt submissions on each problem. Error solutions were only counted as “common” if the submission states were identical to each other and submitted by five or more participants.

\subsubsection{Edit distance to submitted solution}
To analyze how participants' solutions evolved over the course of the task, we looked at the distance between grid state and submitted test output state, which we called ‘edit distance’. We did this for each edit in the full ordered edit sequence for each participant’s first attempt on each problem, allowing us to observe the evolution of participants’ solutions. The distance was measured using the Levenshtein edit distance, which quantifies the number of insertions, deletions, or substitutions required to transform a participant’s current solution state into the target output state at each step (plotted in Figures 10C and 11C) \cite{levenshtein_binary_1966}.

\subsubsection{Edit trajectory efficiency}
To capture efficiency in participants’ problem-solving, we calculated a measure of normalized extra steps. For each participant’s first attempt on a given problem, we counted the number of edit actions taken before the first submission. We then compared this to the minimum number of edits theoretically required to reach the participant’s own submitted solution. The minimum was defined as the smaller of two values: the edit distance between (i) the input grid and the submitted output, and (ii) a blank (all-zero) grid and the submitted output. The normalized extra steps score was computed as the ratio of the participant’s actual number of edits to this minimum. A score of 1 indicates that the participant reached their submitted solution in the minimal number of edits possible, whereas higher values reflect increasing amounts of redundancy, motor error, or exploratory edits. This metric allows us to evaluate editing efficiency across participants, independent of the correctness of the final submission (plotted in Figures 10D and 11D).

\section{Results}
We report participants’ performance on the CogARC task across two experiments and explore how behavior varied across problems, over time, and between solution strategies. Participants attempted to solve up to 75 visual abstract reasoning problems within two hours (same across experiments 1 and 2), each requiring them to infer a transformation rule from input-output examples and generate a corresponding output for a novel test input. We captured participants’ behavior at high temporal resolution, including their sequence of edits to the output grid, viewing patterns, and submission attempts. To assess performance we analyzed task accuracy, problem difficulty score, deliberation time, and edit sequence similarity, and characterized submitted participant solutions and the trajectories leading to those solutions.

\subsection{Experiment 1}
\subsubsection{Validating CogARC as a Tool for Abstract Reasoning}
We first conducted a remote-supervised pilot ($n=40$)(Experiment 1) to validate CogARC as a test of human abstract reasoning (Figure 4). Participants were generally successful across problems ($mean = 89.5\%$ problems correctly solved across all attempts, $SD = 10.2\%$), but this accuracy masked considerable variability in problem difficulty (Figure 4A-C). First-attempt accuracy varied widely across problems. While 76.0\% of tasks were solved correctly on the first attempt by at least half of participants, only 37.3\% exceeded a 75\% success rate and just 8.0\% exceeded 90\%, with some problems solved correctly by almost no participants. Median first-attempt success rate across problems was 69.4\%. Mean deliberation times per problem (the latency between trial onset and a participant’s first edit) were relatively long on average ($mean = 22.3 s$, $SD = 13.5 s$, $range = 68.6 s$), indicating that participants typically spent several seconds inspecting the examples before initiating edits (Figure 4B). Even among the 24 problems that participants solved with 100\% accuracy, there was substantial variability in deliberation times ($mean = 15.6 s$, $SD = 8.6 s$, $range = 44.1 s$).

Given this variability, to better capture problem difficulty, we computed a difficulty score based on the number of attempts required to reach a correct solution (1 = solved on first attempt; 4 = never solved). Across problems, mean difficulty scores ranged widely (Figure 4C). Importantly, mean deliberation time was positively correlated across problems with the mean difficulty score ($r = 0.52$, $p < .001$; Figure 4D), suggesting that participants spent more time examining the examples when the underlying rule was harder to infer.

We next asked whether problem difficulty was systematically related to task properties. No significant differences were observed across core knowledge categories (Figure 4F). In contrast, difficulty increased with experimenter-rated rule complexity, with significant differences between problems rated as 1 vs. 2 ($p < .01$, FDR corrected) and 1 vs. 3 ($p < .01$, FDR corrected) (Figure 4G).

Together, these findings validate CogARC as a sensitive measure of human abstract reasoning. Even in a supervised setting with high overall accuracy, problems varied widely in difficulty and elicited long deliberation times, with performance systematically shaped by problem structure. These pilot results prompted a scaled up, fully remote, large-sample experiment (Experiment 2).

\begin{figure}[H]
    \centering
        \includegraphics[scale=0.8]{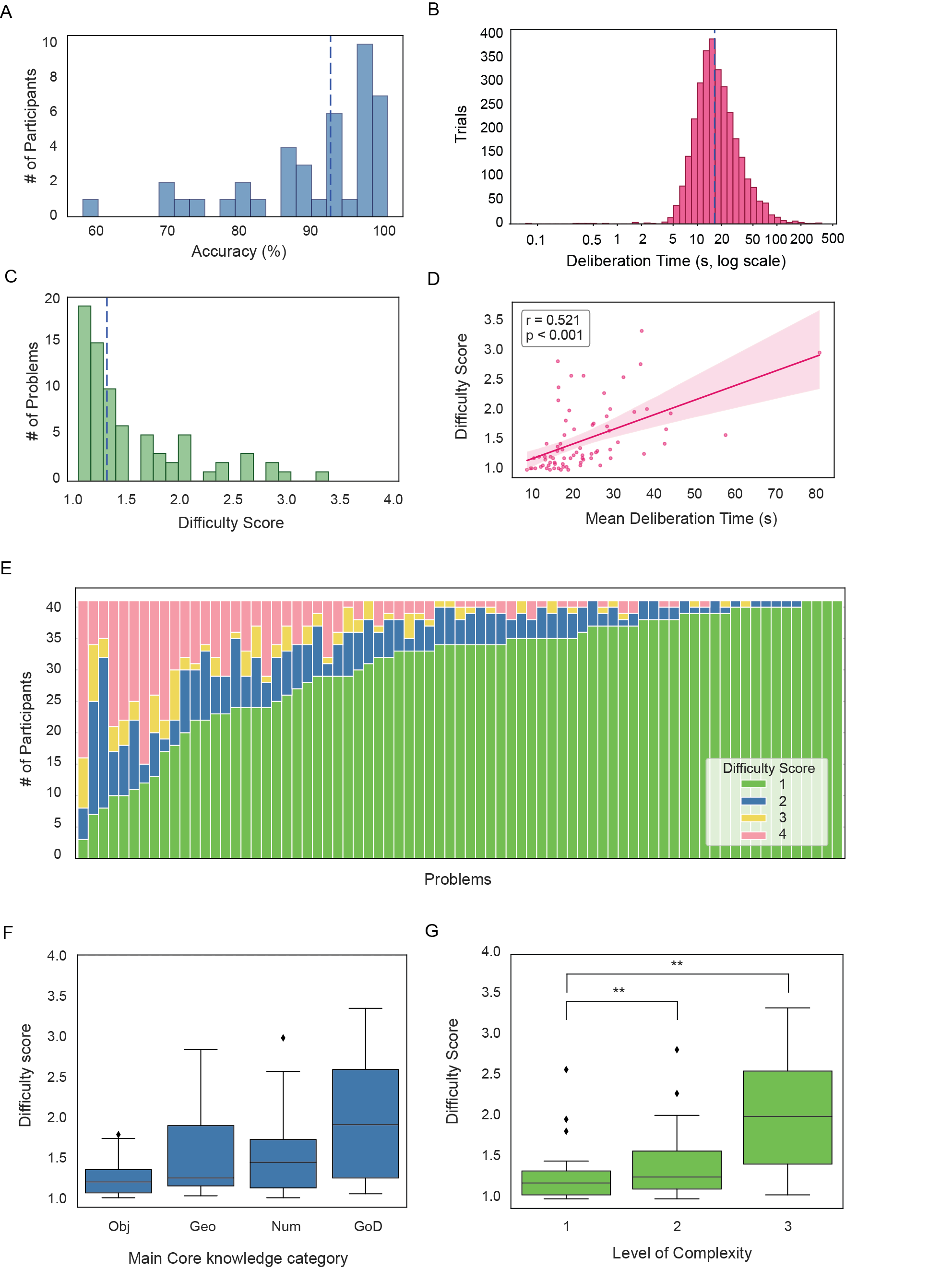}
    \caption{Exploratory results from Experiment 1. A) Distribution of participant accuracy across all problems and attempts ($mean = 89.5\%$, $SD = 10.2\%$). Dashed line indicates the median. B) Distribution of deliberation times (latency from trial onset to first edit; $mean = 22.3 s$, $SD = 13.5 s$). Values are plotted on a $log_{10}$ scale for visualization. Dashed line indicates the median. C) Distribution of mean difficulty scores across problems ($mean = 1.50$, $SD = 0.55$). Dashed line indicates the median. D) Positive correlation between mean deliberation time and mean difficulty score per problem (Pearson $r = 0.52$, $p < .001$). Line reflects best linear fit with 95\% confidence interval indicated by the shaded region. E) Stacked bar plot showing the distribution of difficulty scores across all problems, ordered by number of first attempt successes. F) Comparison of mean difficulty scores per problem across core knowledge categories. (Obj = Objectness, Geo = Geometry and Pattern, Num = Numbers and Counting, GoD = Goal-directedness) G) Comparison of mean difficulty scores across experimenter-assigned complexity levels. (** $p<0.01$)}
    \label{fig:4}
\end{figure}

\subsection{Experiment 1}
\subsubsection{Performance variability reflects rule learning difficulty}

\begin{figure}[H]
    \centering
        \includegraphics[scale=1.0]{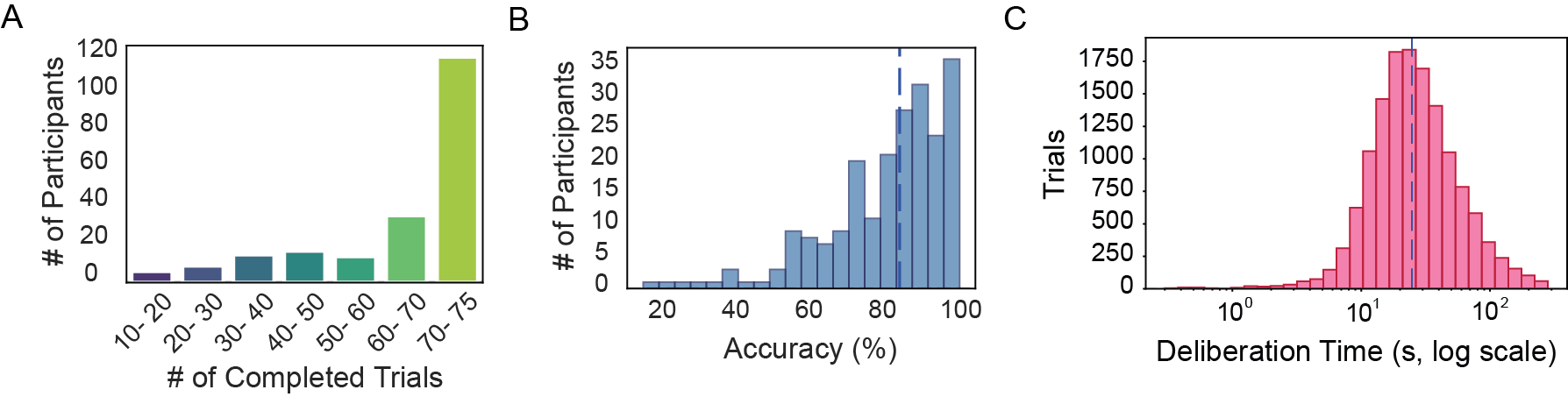}
    \caption{Participant performance across trials. A) Distribution of number of trials completed per participant. B) Histogram showing the distribution of participant accuracy (percentage of correct trials over all trials; $mean = 80.1\%$, $SD = 16.6\%$). Dashed line shows the median. C) Distribution of deliberation times (latency from trial onset to first edit; $mean = 52.74 s$, $median = 25.0 s$, $SD = 1336.9 s$) shown on a log scale for visualization. The top 1\% of values were excluded to improve readability; dashed line indicates the median.}
    \label{fig:5}
\end{figure}

Having established CogARC as a useful measure of abstract reasoning in Experiment 1, we next examined performance in a larger, fully remote sample ($n = 220$). We first examined overall performance on CogARC by analyzing accuracy and deliberation times across the task (Figure 5). Accuracy was defined as the percentage of problems each participant solved correctly across all three attempts. Most participants (55.5\%) completed at least 70 of the 75 trials (Figure 5A). On average, participants solved 58.2 problems ($median = 70$). Humans were generally proficient at ARC, with a mean accuracy of 80.1\% ($SD = 16.6\%$) across attempts, a median of 83.6\%, and a range from 13.7\% to 100\% (Figure 5B).

These results are broadly consistent with prior large-scale human benchmarks on ARC. such For example, H-ARC reported an average task accuracy of 76.2\% across training-set problems after up to three attempts, with mean first-attempt accuracy of 59.9\% and second-attempt accuracy of 72.6\%. H-ARC also found substantial heterogeneity across tasks, including a subset that was universally solvable and a small number that no participants could solve \cite{legris_comprehensive_2025}. Together, these convergent findings suggest that CogARC elicits human performance patterns comparable to those observed in prior ARC studies.

As in Experiment 1, first-attempt success rates varied substantially across tasks. Although a majority of problems were solved correctly on the first attempt by most participants, relatively few reached near-ceiling performance, and some elicited consistently low first-attempt success. Specifically, 78.7\% of tasks were solved on the first attempt by at least half of participants, 46.7\% by at least 75\%, and 17.3\% by at least 90\%, with first-attempt success rates ranging from 12.3\% to 98.4\% across problems ($median = 73.9\%$). This pattern closely mirrors the distribution observed in Experiment 1, suggesting that CogARC reliably captures problem-level variation in abstract reasoning difficulty across experimental contexts.

Deliberation times in CogARC were broadly distributed and positively skewed, with a long right tail characteristic of response time data (Figure 5C). For visualization, the top 1\% of values were clipped, but the full distribution was used in all statistical analyses.

To capture differences in task difficulty more precisely, we once again used difficulty score (1 = success on first attempt; 4 = no success across all three attempts). Across problems, the mean difficulty score was 1.62 ($SD = 0.57$). On average, accuracy increased by 13.3\% from the first to the second attempt and by an additional 4.0\% on the third attempt.

Despite high overall accuracy, difficulty varied substantially across the 75 problems (Figure 6A).   We explored what factors best explained the variation in difficulty score. Harder problems elicited longer deliberation times ($r = 0.69$, $p < .001$), suggesting that participants spent more time inspecting the examples before starting to edit on their first attempt. To capture how strategies differed across participants, we analyzed edit sequences, defined as the ordered set of (x, y, color) changes participants made to the output grid during each attempt. We then computed edit similarity using the Jaccard similarity index across participants’ edit sequences for each problem. Higher values indicate more convergence in strategies, while lower values indicate divergence. Difficulty was strongly negatively correlated with edit similarity ($r = –0.83$, $p < .001$; Figure 6C), showing that participants’ approaches were more consistent on easier problems and more varied on harder ones.

We compared mean difficulty scores with the independent experimenter-assigned complexity ratings (1–3), which reflected the number and type of rule components required. As in Experiment 1, the two measures were strongly correlated (Pearson $r = 0.61$, $p < .001$; Figure 6C), supporting the interpretability of difficulty score as a behavioral correlate of abstract rule complexity.

Difficulty showed no significant relationship with normalized extra steps in edit sequences ($r = –0.01$, $p = .93$). This indicates that efficiency during the act of solving (how directly participants moved from input to output once they began editing) did not predict success. In other words, the time participants spent reasoning before starting to edit was strongly associated with outcomes, but their efficiency while editing was not. Finally, difficulty also showed no significant relationship with low-level perceptual features such as test grid size ($r = 0.05$, $p = 0.65$), number of colors ($r = –0.05$, $p = 0.66$), or number of edits required to reach the test output ($r = 0.12$, $p = 0.30$). Together, these findings indicate that variability in performance on CogARC primarily reflected differences in rule inference and learning, rather than perceptual or motor demands.

\begin{figure}[H]
    \centering
        \includegraphics[scale=1.0]{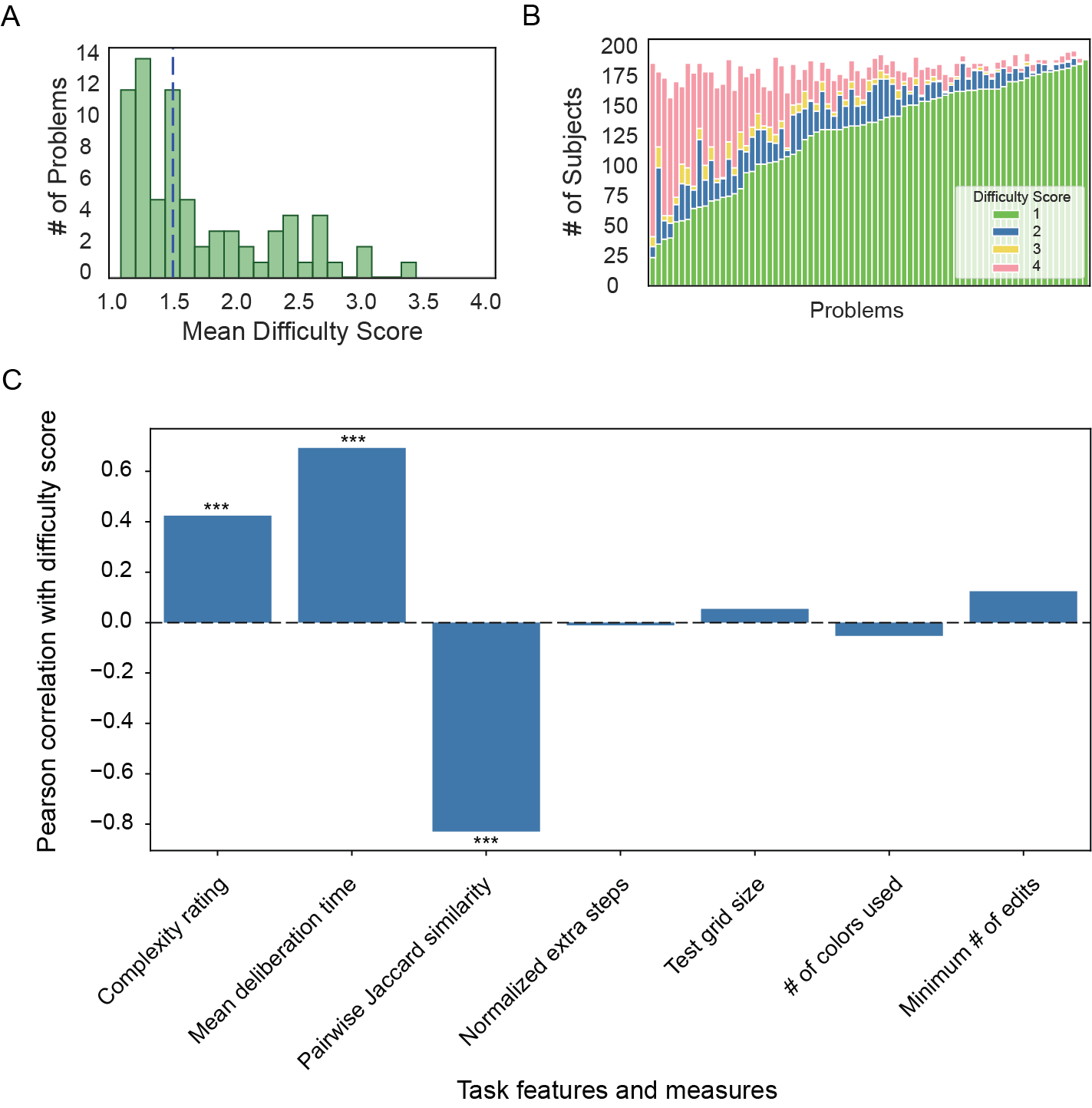}
    \caption{Difficulty across problems and correlation with task features and other measures of performance. A) Distribution of mean difficulty scores for each problem. Dashed line shows the median. B) Stacked bar plot showing the distribution of difficulty scores per problem, ordered by number of first attempt successes. C) Comparison of Pearson correlation coefficients between mean difficulty score per problem and different task features and measures. *** denotes $p<0.001$.}
    \label{fig:6}
\end{figure}

\subsubsection{Over time, participants get faster but not better at learning the rules}

\begin{figure}[H]
    \centering
        \includegraphics[scale=1.0]{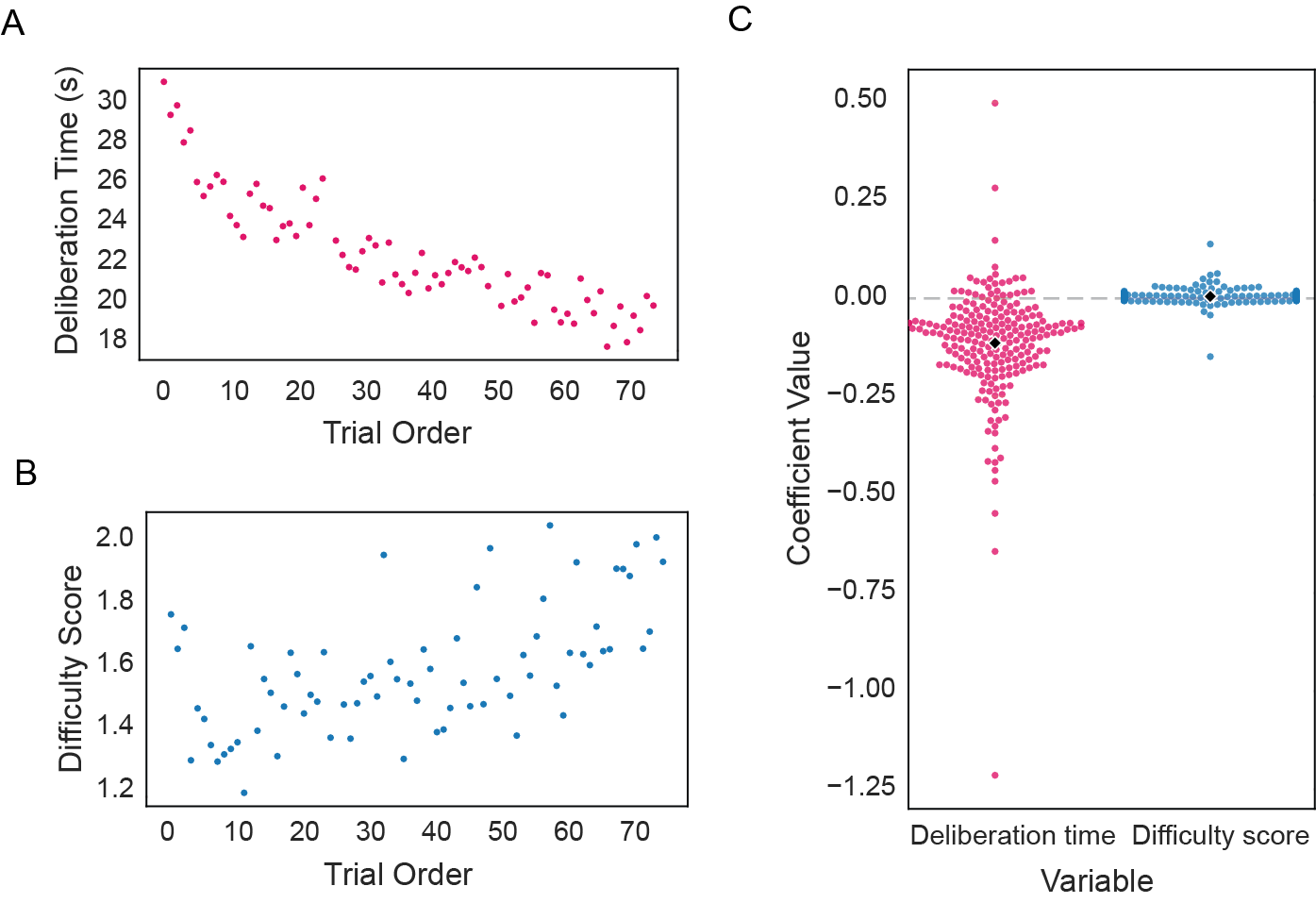}
    \caption{Performance over time. A) Mean deliberation time per trial over the course of 75 trials. B) Mean difficulty score per trial. C) Deliberation time and difficulty score regression coefficient values for each participant over time. Black diamonds show the mean. The dashed line is at 0.}
    \label{fig:7}
\end{figure}

Participants were given 2 hours to solve as many problems as they could out of 75 problems, presented in a random order. To assess changes in behavior over time, we computed individual linear regressions predicting deliberation time and difficulty score from trial number.

The average slope was significantly negative ($mean= -0.114$, $t(217) = -11.85$, $p < .001$), indicating that participants grew faster at initiating responses with exposure to the task (Figure 7A, 7C). However, this increased efficiency did not correspond to improved task performance: average regression slopes for difficulty score were slightly positive ($mean=0.005$, $t(217) = 3.98$, $p < .001$) (Figure 7B, 7C). These results suggest that participants were consistently getting faster but less accurate at solving ARC problems over time.

To test whether these changes were related, we examined the relationship between participants’ deliberation and difficulty slopes. After winsorizing outliers beyond the 1st and 99th percentiles (removing 6 total values, 2.7\% of the sample), the correlation was weak and nonsignificant ($r = -0.12$, $p = .066$). This pattern suggests that, although participants as a group became faster and slightly less accurate, these trends were largely independent across individuals. In other words, participants who sped up the most were not necessarily those who lost the most accuracy.

Together, these findings indicate that participants became faster at initiating responses as the task progressed, but this increased speed did not consistently trade off with performance. Instead, the pattern may reflect growing task fluency, reduced deliberation, or mild cognitive fatigue over time, rather than a strict speed-accuracy tradeoff. In keeping with the interpretation of ARC and RPM-style tasks as demanding fluid reasoning rather than incremental skill learning, there was no evidence that practice from earlier problems benefited accuracy on later problems.

\subsubsection{Common errors suggest shared biases in rule learning}

\begin{figure}[H]
    \centering
        \includegraphics[scale=1.0]{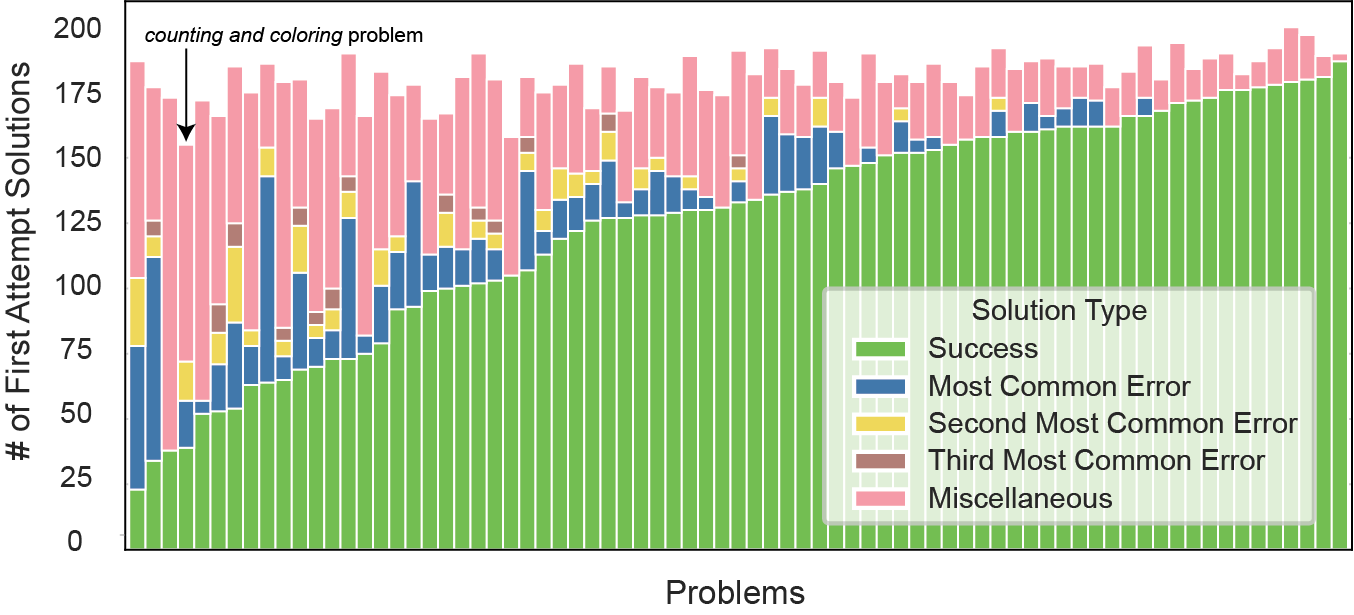}
    \caption{Percentage of first attempt solutions per problem. The stacked bars show the percentage of submitted solution types (success, common errors, and miscellaneous) for the first attempt on each problem, ordered by number of first attempt successes. Some problems have a higher proportion of common error solutions than others.}
    \label{fig:8}
\end{figure}

To examine whether participants converged on the same types of errors, we categorized first-attempt submissions into correct solutions and up to three common incorrect solutions per problem (Figure 8). Common errors were defined as identical outputs submitted by at least five participants. As with difficulty scores, the prevalence of common errors varied across problems. Some problems elicited disproportionately high rates of shared errors, suggesting that the input–output examples supported an alternative but incorrect generalization. These common errors were also observed in Experiment 1 (Supp. Figure 1), indicating that such behaviors were consistent across experimental contexts.

To better understand how participants arrived at these solutions, we analyzed edit trajectories, defined as the evolving distance between a participant’s grid state and the correct solution at each step (Figure 9). We also measured the amount of extra edit actions that participants took required to reach their submitted solution (number of actions taken divided by the minimum required), to provide a measure of how direct their trajectories were (see \textit{normalized extra steps} in Methods). These trajectories revealed strategy dynamics not apparent from final outputs alone.

\begin{figure}
  \begin{minipage}[c]{0.52\textwidth}
    \includegraphics[width=\textwidth]{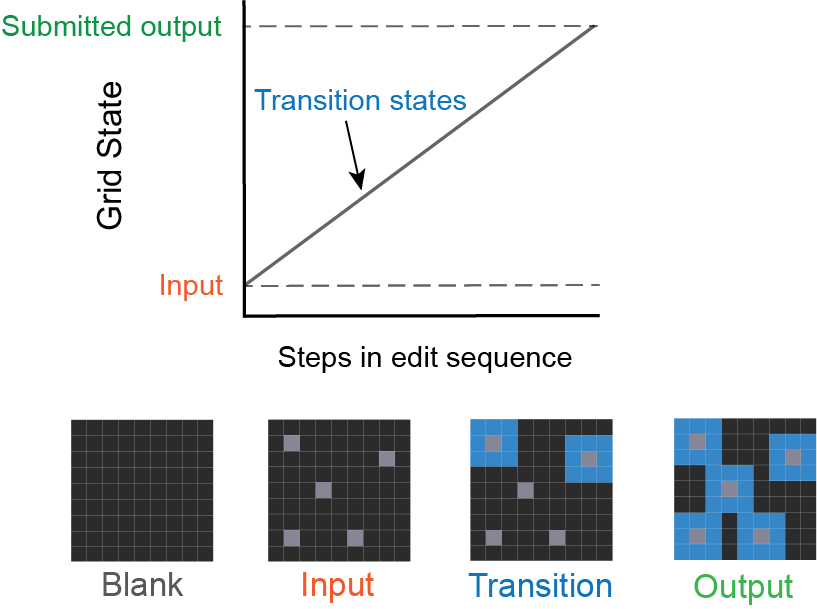}
  \end{minipage}\hfill
  \begin{minipage}[c]{0.43\textwidth}
    \caption{
       Schematic of solution trajectory plots. The solution trajectory plots show how close the grid state is to the test output at any step of the edit sequence. The y-axis values are the normalized distance to the submitted output grid, which are calculated by the number of tiles in the grid that exactly match the tiles of the submitted output in color and location. Edit sequences almost all begin with copying from input (“input”), and proceed through a series of transition states to reach the submitted output. This output can match the correct solution, or fall short.
    } \label{fig:9}
  \end{minipage}
\end{figure}

For example, in the counting-and-coloring problem (Figure 10), trajectories were generally short and direct, with most participants converging quickly on either the correct solution or one of the two most common errors. Notably, the distribution of normalized extra steps for Top Error 2 was mostly skewed to the right but with a smaller second mode toward the tail, which suggests most participants moved directly to the outcome but a smaller subset initially explored an alternative path before redoing their work and converging on the same incorrect solution.

In contrast, the pattern-completion problem (Figure 11) elicited more variable trajectories across solution types. Paths were longer and more heterogeneous, with many participants showing diffuse or plateauing trajectories before submitting. As in the counting problem, Top Error 2 displayed a right skewed distribution of normalized extra steps with a noticeable bump at the tail, whereas the other solution types showed broader distributions.

These contrasts suggest that problem type influences not only accuracy but also the dynamics of solution strategies, with some errors emerging from direct application of a misinterpreted rule and others from more exploratory or uncertain search.

\begin{figure}
  \begin{minipage}[c]{0.63\textwidth}
    \includegraphics[width=\textwidth]{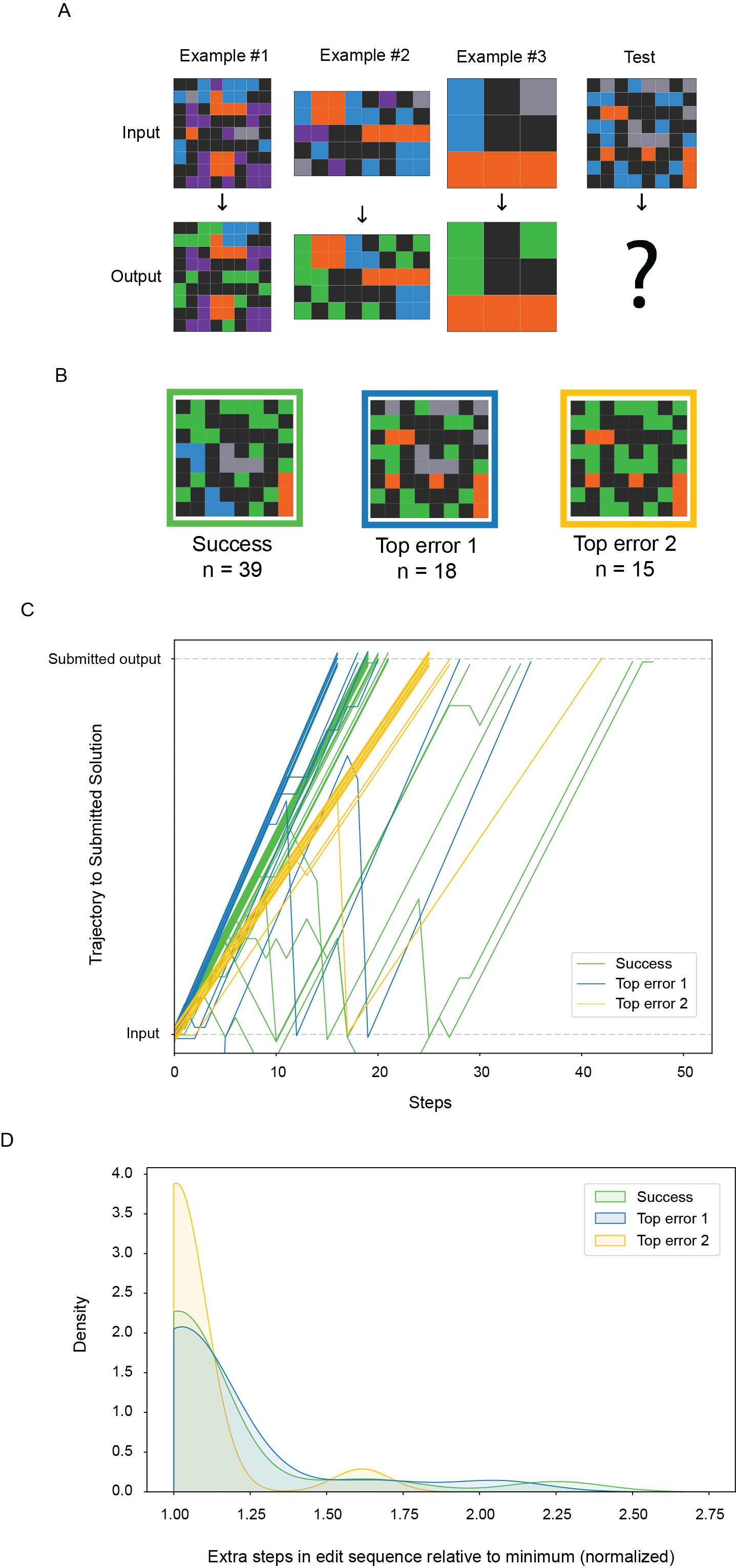}
  \end{minipage}\hfill
  \begin{minipage}[c]{0.31\textwidth}
    \caption{
       Efficient convergence on both correct and incorrect solutions in a counting-and-coloring problem. A) Input–output examples and test input. The correct rule required recoloring all connected groups of fewer than three same-colored tiles to green. B) Most common first-attempt outputs. Each output is shown with a colored frame indicating solution type. In addition to the correct solution, two incorrect outputs were repeatedly produced, indicating that failures were concentrated on a small number of systematic alternatives. C) Edit-distance trajectories from input to submitted output. Each line represents one participant’s edit sequence, plotted as normalized distance from their submitted solution at each step, allowing trajectories corresponding to different solution types to be shown on the same y-axis. A fixed vertical jitter was applied so that overlapping trajectories could be visualized. Trajectories for correct and incorrect solutions are short and largely monotonic, showing rapid convergence with little evidence of extended exploration aside from a few participants per solution group who reset their grids mid-trajectory.  D) Distribution of normalized extra steps relative to the minimum possible edit length. Successes and common errors overlap near the minimum, indicating comparable path lengths across solution types. For Top Error 2, a secondary mode in the tail reflects a subset of participants who initially deviated before re-converging on the same incorrect outcome. 
    } \label{fig:10}
  \end{minipage}
\end{figure}
\begin{figure}
  \begin{minipage}[c]{0.63\textwidth}
    \includegraphics[width=\textwidth]{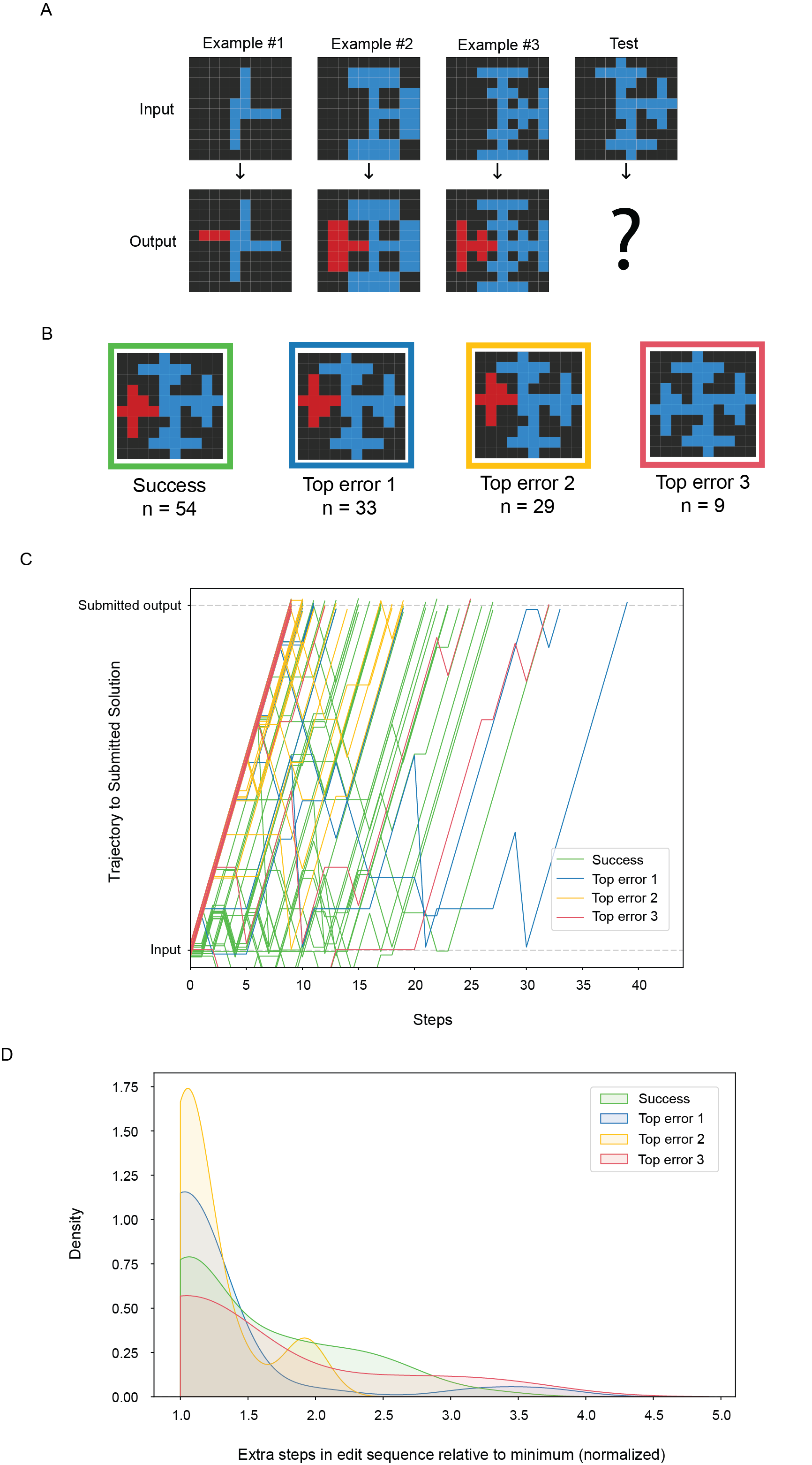}
  \end{minipage}\hfill
  \begin{minipage}[c]{0.31\textwidth}
    \caption{
       Shared incorrect solutions despite heterogeneous edit trajectories in a pattern-completion problem. A) Input–output examples and test input. The correct transformation required completing a rotational pattern by adding the red segment. B) Most common first-attempt outputs. Each output is shown with a colored frame indicating solution type. Although the correct solution was produced by 54 participants, three distinct incorrect outputs were also frequent and collectively more common, reflecting consistent misgeneralizations of the transformation rule. C) Edit-distance trajectories from input to submitted output. Each line represents one participant’s edit sequence, plotted as normalized distance from their submitted solution at each step, allowing trajectories corresponding to different solution types to be shown on the same y-axis. A fixed vertical jitter was applied so that overlapping trajectories could be visualized. Compared to Figure 10, trajectories are more variable across all solution types, with many participants exhibiting irregular paths prior to submission. D) Distribution of normalized extra steps relative to the minimum possible edit length. While all solution types include near-minimal sequences, incorrect solutions show broader and more right-skewed distributions, including a secondary mode for Top Error 2, consistent with partial exploration followed by convergence on a shared incorrect solution. 
    } \label{fig:11}
  \end{minipage}
\end{figure}

\section{Discussion and Conclusions}
We developed the CogARC paradigm so we could examine how people infer and apply abstract rules from minimal examples. Performance varied widely across problems and was highly related to rule complexity and how long people deliberated before starting to solve. Thus, even though the ARC is presented as a challenge for artificial general intelligence that humans can perform, we found that the human performance on the task was highly variable across individuals. Over time, participants responded more quickly but became slightly less accurate. Even when incorrect, participants often converged on the same solutions and edit sequences, revealing shared errors. These patterns may reflect inductive biases, the systematic tendencies that guide how people generalize from sparse data \cite{griffiths_probabilistic_2010, tenenbaum_how_2011, lake_building_2017}. Inductive biases are not flaws but rather essential constraints: they enable people to learn quickly from limited information by narrowing the hypothesis space, even if this means they sometimes lead to systematic errors.

\subsection{Rule complexity contributes to performance variability}
Participants were generally successful at solving the tasks, but accuracy varied substantially across problems, with overall success rates similar to those reported previously for the H-ARC paradigm \cite{legris_h-arc_2024, legris_comprehensive_2025}. This variation was strongly predicted by researcher-rated rule complexity, suggesting that differences in performance reflected challenges in rule inference. These results validate CogARC as a benchmark for studying reasoning across a structured set of abstract transformations. They also align with findings from ConceptARC, where problems grouped by rule type elicited consistent differences in human accuracy \cite{moskvichev_conceptarc_2023}. Together, these results suggest that task difficulty in ARC-style benchmarks is shaped primarily by conceptual structure such as the number of transformations or abstract concepts involved rather than low-level features of the visual input.

\subsection{Efficiency gains over time}
Participants became consistently faster to initiate solutions as the task progressed, whereas task accuracy showed a small but steady decline. This pattern might superficially suggest a speed-accuracy \cite{heitz_speed-accuracy_2014} tradeoff, where participants respond more quickly at the expense of performance. However, our individual-level analyses did not support a direct link between these trends: the rate of decrease in deliberation time (i.e., slope across trials) was not significantly correlated with the rate of increase in difficulty scores. Thus, while participants as a group became faster and slightly less accurate over time, individuals who sped up the most were not necessarily those whose performance declined most strongly.

Together, these results suggest that faster responding reflects general task adaptation (such as increasing fluency with the interface or rule format) rather than an explicit tradeoff between speed and accuracy. At the same time, it remains possible that fatigue or reduced motivation contributed to the slight decline in accuracy over the course of the session. Because participants completed up to 75 problems over two hours, later trials may have elicited less sustained attention, leading to faster but less deliberative responses.

Prior work supports the idea that these trends are more likely to reflect changes in task familiarity and general engagement rather than changes in underlying reasoning capacity per se. Fluid intelligence, or the ability to solve novel problems independent of prior knowledge, is generally considered relatively stable in adulthood and resistant to short-term training effects \cite{cattell_theory_1963, engle_working_1999}. While performance on working memory tasks can improve with practice, evidence that such training leads to reliable gains in fluid reasoning has been mixed and difficulty to replicate. Studies have shown that improvements following working memory training tend to be task-specific, with limited transfer to other measures of cognitive ability \cite{redick_no_2013}. This pattern in consistent with our findings in CogARC: although participants became faster over time ,there was no evidence of improved rule induction ability with practice, suggesting that increased efficiency likely reflects task adaptation rather than enhanced abstract reasoning.

Research on intuitive reasoning further suggests that people are faster and more accurate when problems align with their expectations or allow for heuristic shortcuts \cite{babai_intuitive_2006, gillard_proportional_2009}. In our data, faster responses may reflect strategies that reduce effort without guaranteeing correct generalization, such as internalized heuristics or increased pattern recognition. Supporting this view, Goldhammer and Entink (2011) found that executive attention, rather than perceptual speed, predicted performance on Raven’s Advanced Progressive Matrices \cite{goldhammer_speed_2011}. Participants who could efficiently validate mental models tended to solve problems faster and more accurately, which is consistent with the idea that CogARC participants learned to constrain the solution space more effectively over time. This pattern also aligns with findings from H-ARC, where participants spent significantly less time on training-set problems than on harder evaluation-set ones, despite similar task formats \cite{legris_comprehensive_2025}.

\subsection{Structured errors reveal shared strategies}
ARC was originally designed around the idea that people bring conceptual priors to reasoning tasks. Its rule space was explicitly inspired by core knowledge theory \cite{chollet_measure_2019}, which proposes that humans possess domain-specific systems for reasoning about objects, agents, number, and geometry \cite{spelke_core_2007, carey_origin_2009}. These priors can be thought of as developmentally early-emerging inductive biases that constrain the hypothesis space and enable fast learning from sparse data. Although CogARC does not directly measure these systems, our results suggest that such structured knowledge may influence how participants generalize from limited input–output examples.

Because the task is open-ended and permits a wide range of responses, one might expect high variability in incorrect outputs. Instead, we observed the opposite: errors were highly structured and interpretable. In several problems, numerous participants who failed to solve the task still produced the same incorrect output or followed similar edit trajectories. These shared “wrong answers” suggest that people were guided by similar heuristics or inductive biases about how transformations should work, even when those assumptions were not strictly consistent with the transformations shown in the example grids. For example, in the pattern-completion problem, the combined number of participants who made common errors outnumbered those who submitted the correct one ($n=71$ vs. $n=54$) which indicates systematic misgeneralizations based on plausible but incorrect extrapolation of the transformation rule (Figure 11).

Process-level analyses confirmed that errors were not random. In both the counting-and-coloring and pattern-completion problems (Figures 10-11), participants who converged on common errors tended to follow consistent trajectory shapes, steadily reducing distance from the submitted output before reaching an incorrect solution. This indicates that even incorrect rules were applied systematically and efficiently, rather than arising from noise or disengagement. Consistent with this, distributions of edit sequence lengths for successes and common errors largely overlapped near the minimum, showing that many incorrect solutions were reached with the same efficiency as correct ones. In some cases, such as Top Error 2 for both tasks, the distribution was bimodal: most participants went directly to the mistaken solution, while a smaller subset appeared to “redo” their work and still converged on the same error. These patterns reinforce our finding that efficiency during solving is not diagnostic of correctness, and that shared strategies underlie both successes and failures.

Together, these results suggest that inductive biases play a dual role in CogARC. They support rapid generalization from sparse examples, but can also give rise to consistent misgeneralizations when intuitive assumptions conflict with the ground truth. Such patterns mirror prior findings from ConceptARC, where human errors were often interpretable and related to the target rule, reflecting partial understanding rather than arbitrary failure \cite{moskvichev_conceptarc_2023}. Similarly, H-ARC reported that easier problems elicited more convergent error patterns, consistent with the idea that shared inductive biases shape both successful and unsuccessful reasoning in ARC-style tasks \cite{legris_comprehensive_2025}. These error patterns are not merely byproducts of failure, but reflect the underlying hypothesis space that participants explore during abstract reasoning.

Recent computational work by Do et al. (2025) has shown that explicitly incorporating such human error structure into model design can yield systems that reproduce human-like patterns of success and failure on ARC problems \cite{do_priors_2025}. The model architecture in Anonymous et al. (2025) was informed by structured error patterns identified in CogARC behavioral data, using these misgeneralizations to constrain how abstract representations were constructed and replicate human errors. This suggests that human errors can serve as a valuable signal for constraining abstract representations in ways that improve model interpretability.

\subsection{Implications for AI Evaluation and Modeling Human Reasoning}
These findings have direct implications for how abstract reasoning is evaluated and modeled in artificial systems, particularly when the goal is to capture human-like reasoning rather than improving task-level accuracy alone. When humans fail on ARC problems, their errors can often be traced back to plausible hypotheses about the underlying rule. Such failures reflect partial understanding, overgeneralization, or the application of intuitive heuristics, rather than random or unconstrained behavior \cite{tversky_judgment_1974, evans_heuristic_1984, schunn_generalityspecificity_1999, rozenblit_misunderstood_2002}. Critically, this structure makes human error informative: they can reveal the inductive biases and representational assumptions that guide reasoning, insights that cannot be recovered from binary accuracy measures alone \cite{tenenbaum_how_2011, ullman_mind_2017}.

In contrast, many contemporary AI systems, particularly large language models and program synthesis approaches, offer limited visibility into how solutions are generated or why failures occur \cite{lake_building_2017, binz_using_2023, turpin_language_2023, lanham_measuring_2023, mitchell_artificial_2025}. As a result, failures in these systems are often difficult to diagnose or interpret, even when performance appears strong. This gap highlights the value of behavioral tasks like CogARC, which capture not only outcomes but the strategies and trajectories that lead up to them. Such rich data afford us a unique chance to reverse-engineer the latent structures behind human reasoning \cite{tenenbaum_how_2011, lake_human-level_2015}.

This process-level view is increasingly relevant as large-scale foundation models are being repurposed to simulate human behavior. Recent work has shown that models such as Centaur, a fine-tuned variant of Meta’s LLM, can predict individual human responses across a range of cognitive tasks, revealing representational similarities between neural and model embeddings \cite{binz_foundation_2025}. Such models provide a powerful framework for testing hypotheses about knowledge representation and information processing in humans, rather than merely achieving performance thresholds. CogARC contributes complementary evidence to this effort: it captures how people explore, generalize, and misgeneralize abstract rules, offering empirical constraints that could guide model training and interpretation.

From an AI-engineering perspective, this has important implications. A recent study from Apple tested frontier reasoning models such as OpenAI’s o-series and Anthropic’s Claude variants on logic puzzles like the Tower of Hanoi. The results revealed fundamental limitations in AI “reasoning”: large reasoning models (LRMs) tended to “overthink” on simpler problems and exhibited complete failure when task complexity exceeded a threshold \cite{shojaee_illusion_2025}. Moreover, similar work testing LLMs on ARC shows that even with large computational resources and chain-of-thought prompting, models can fail to maintain performance on tasks requiring structured planning \cite{gendron_large_2023, wu_understanding_2025}. These systems are powerful, but their reasoning remains computationally intensive and difficult to interpret, whereas human solvers can infer novel rules from a handful of examples and often verbalize intuitive and coherent (if imperfect) explanations of their reasoning \cite{acquaviva_communicating_2021}. This discrepancy raises an important question: what would it mean for AI to truly “think” like a human? Matching human accuracy is not enough. Models of human cognition must also replicate the structure of human reasoning, including its biases, error patterns, and computational efficiency, if they are to generate explanations that are intuitive and relatable to human users. Our findings suggest that incorporating cognitive constraints (e.g., inductive biases, limited memory and training data, attention bottlenecks, heuristic search) may be key to building AI systems that are not only more interpretable, but also more aligned with human reasoning processes \cite{hosseini_artificial_2024}.

Progress on ARC benchmarks has been rapid — rising from ~20\% in 2019 to over 50\% in the 2024 contest \cite{mitchell_abstraction_2021, chollet_arc_2025} — but this growth has come largely from scale, not insight. Models like GPT-4o and Claude 3 can solve many ARC tasks, but their solutions lack the transparency, flexibility, and failure interpretability of human problem solvers \cite{wu_understanding_2025}. This suggests that progress toward human-like AI will increasingly depend on integrating cognition-inspired constraints (including inductive biases, working memory, attentional mechanisms, and heuristic search) into model design. By grounding models in human behavioral data, including their systematic errors, we can develop new approaches to AI that approximate human reasoning.

\subsection{Conclusion}
Our findings highlight that human abstract reasoning in ARC is characterized by both variability and structure. Some solutions and errors diverged widely across participants, but others showed striking convergence, pointing to shared strategies and inductive biases. These structured tendencies suggest that even when people fail, their reasoning is guided by logical constraints rather than random exploration. CogARC provides a framework for studying these dynamics at scale, complementing traditional accuracy-based measures and offering a richer view of the cognitive processes underlying abstraction. Beyond advancing our understanding of human reasoning, these results underscore the potential of behavioral benchmarks like CogARC to inform the development of AI systems that are not only more accurate, but also more interpretable, efficient, and informed by human cognition.


\section*{Acknowledgments}
We thank Jingxuan Guo, Thomas Morin, and Nicholas Wayhs for helpful discussions and assistance. This work was supported by the Office of Naval Research Multidisciplinary University Research Initiative (MURI) grant N00014-16-1-2832, the Office of Naval Research Defense University Research Instrumentation Program (DURIP) grant N00014-17-1-2304, and the Boston University Kilachand Fund. We also thank the study participants for their time and effort.

\section{Ethics Statement}
Human-participant procedures were approved by an Institutional Review Board. Informed consent was obtained from all participants.

\section{Data Availability}
The CogARC behavioral dataset, including the full task set, participant action logs, common solutions, and task metadata, is publicly available at DOI: 10.5281/ZENODO.18177487 under a CC-BY license \cite{ahn_cogarc_2026}. Any additional materials are available from the corresponding author upon reasonable request.

\section{Conflict of Interest}
The authors declare no conflicts of interest.

\medskip

\bibliography{references} 

\clearpage
\appendix
\renewcommand{\thepage}{S\arabic{page}}
\renewcommand{\thesection}{S\arabic{section}}
\renewcommand{\thetable}{S\arabic{table}}
\renewcommand{\thefigure}{S\arabic{figure}}
\renewcommand{\figurename}{Supplementary Figure}
\setcounter{figure}{0}
\setcounter{page}{1}

\section*{Supporting Information}
\begin{center}
    \textbf{Supporting Information for:\\ Exploring Human Behavior during Abstract Rule Inference and Problem Solving with the Cognitive Abstraction and Reasoning Corpus}
\end{center}

\section{Supplementary Material}

\begin{figure}[H]
    \centering
        \includegraphics[scale=1.0]{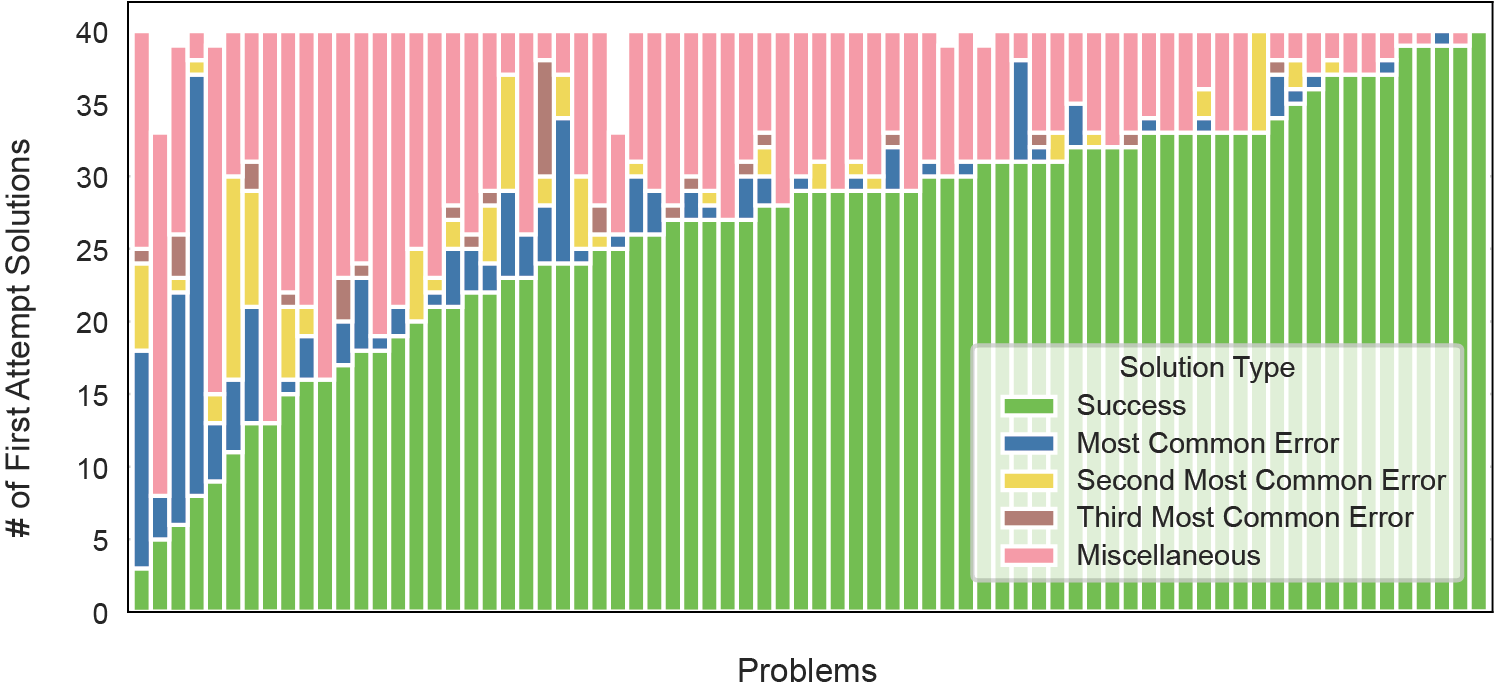}
    \caption{Shared common errors between experiments 1 and 2. We categorized participant first attempt submissions by the common solutions found in experiment 2, and found that the common errors persisted across the different studies.}
    \label{fig:s1}
\end{figure}

\begin{figure}[H]
    \centering
        \includegraphics[scale=0.8]{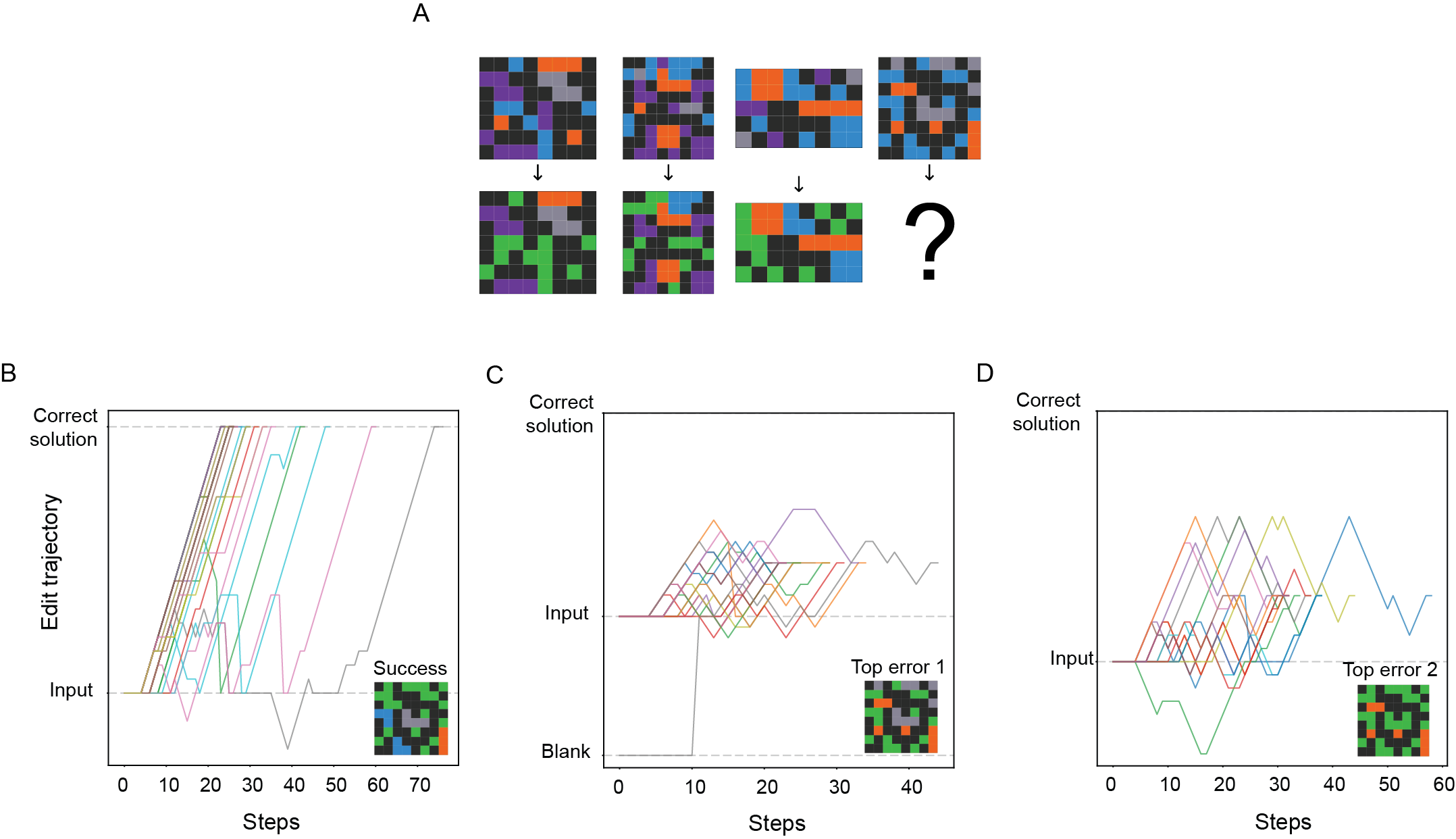}
    \caption{Solution-specific edit trajectories relative to the correct outcome in the counting-and-coloring problem. A) Input–output examples and test input for the counting-and-coloring problem shown in Figure 10. The correct transformation required recoloring all connected groups of fewer than three same-colored tiles to green. B) Edit-distance trajectories shown separately for each solution type: correct solution (left), Top Error 1 (center), and Top Error 2 (right). Each line represents one participant’s edit sequence. Unlike Figure 10, the y-axis here shows unnormalized edit distance to the correct solution, allowing direct comparison of how trajectories associated with different outcomes approach or diverge from the correct target.}
    \label{fig:s4}
\end{figure}

\begin{figure}
  \begin{minipage}[c]{0.63\textwidth}
    \includegraphics[width=\textwidth]{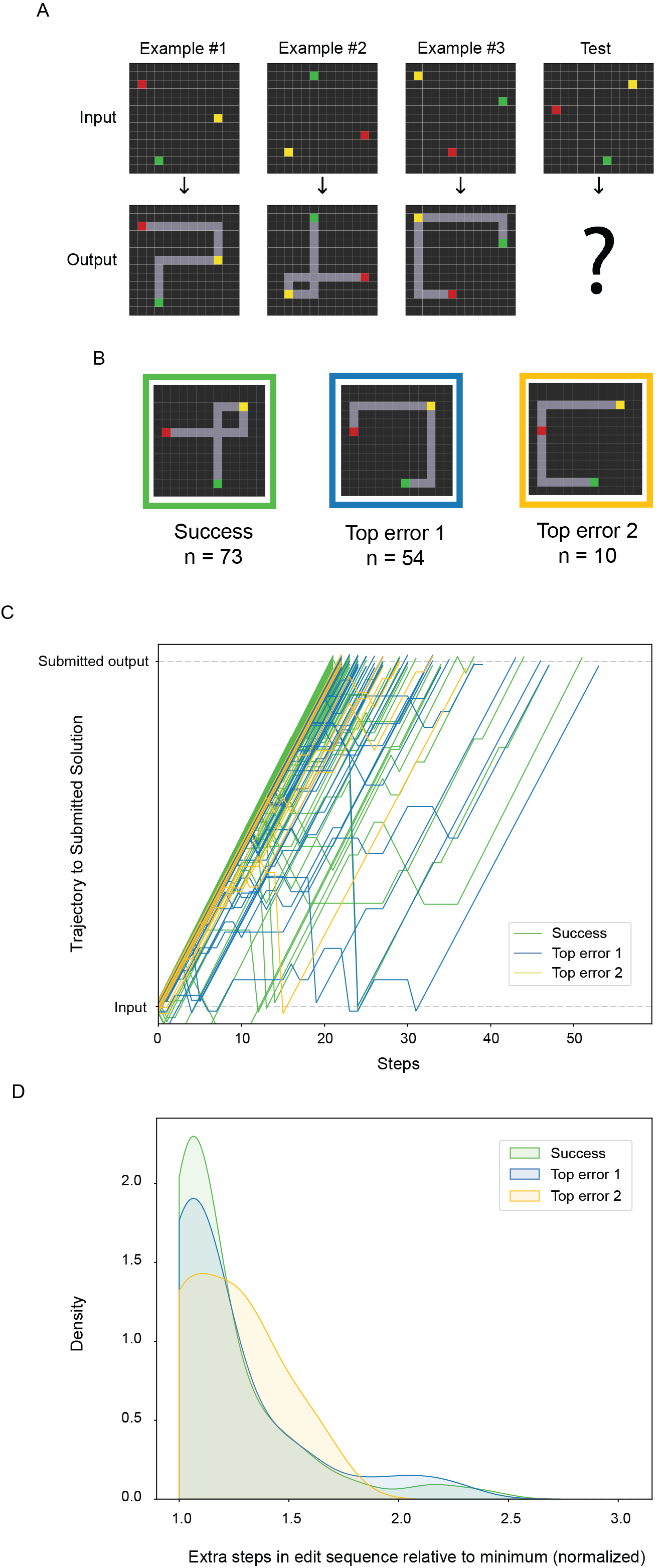}
  \end{minipage}\hfill
  \begin{minipage}[c]{0.31\textwidth}
    \caption{
       Relatively uniform edit efficiency with occasional restarting in a path-connecting problem. A) Input–output examples and test input. The correct transformation required connecting the red, yellow, and green cells by drawing a gray path. B) Most common first-attempt outputs. Each output is shown with a colored frame indicating solution type. In addition to the correct solution, two incorrect outputs were also frequently produced, indicating that errors were concentrated on a small number of alternatives. C) Edit-distance trajectories from input to submitted output. Each line represents one participant’s edit sequence, plotted as normalized distance from their submitted solution at each step, allowing trajectories corresponding to different solution types to be shown on the same y-axis. A fixed vertical jitter was applied so that overlapping trajectories could be visualized. Across solution types, most trajectories are short and directed, with only a small subset of participants exhibiting more indirect paths or grid resets. D) Distribution of normalized extra steps relative to the minimum possible edit length. For both correct and incorrect solutions, most edit sequences cluster near the minimum, indicating broadly similar efficiency across solution types, with a minority of longer sequences reflecting occasional restarting behavior.
    } \label{fig:s2}
  \end{minipage}
\end{figure}

\begin{figure}
  \begin{minipage}[c]{0.69\textwidth}
    \includegraphics[width=\textwidth]{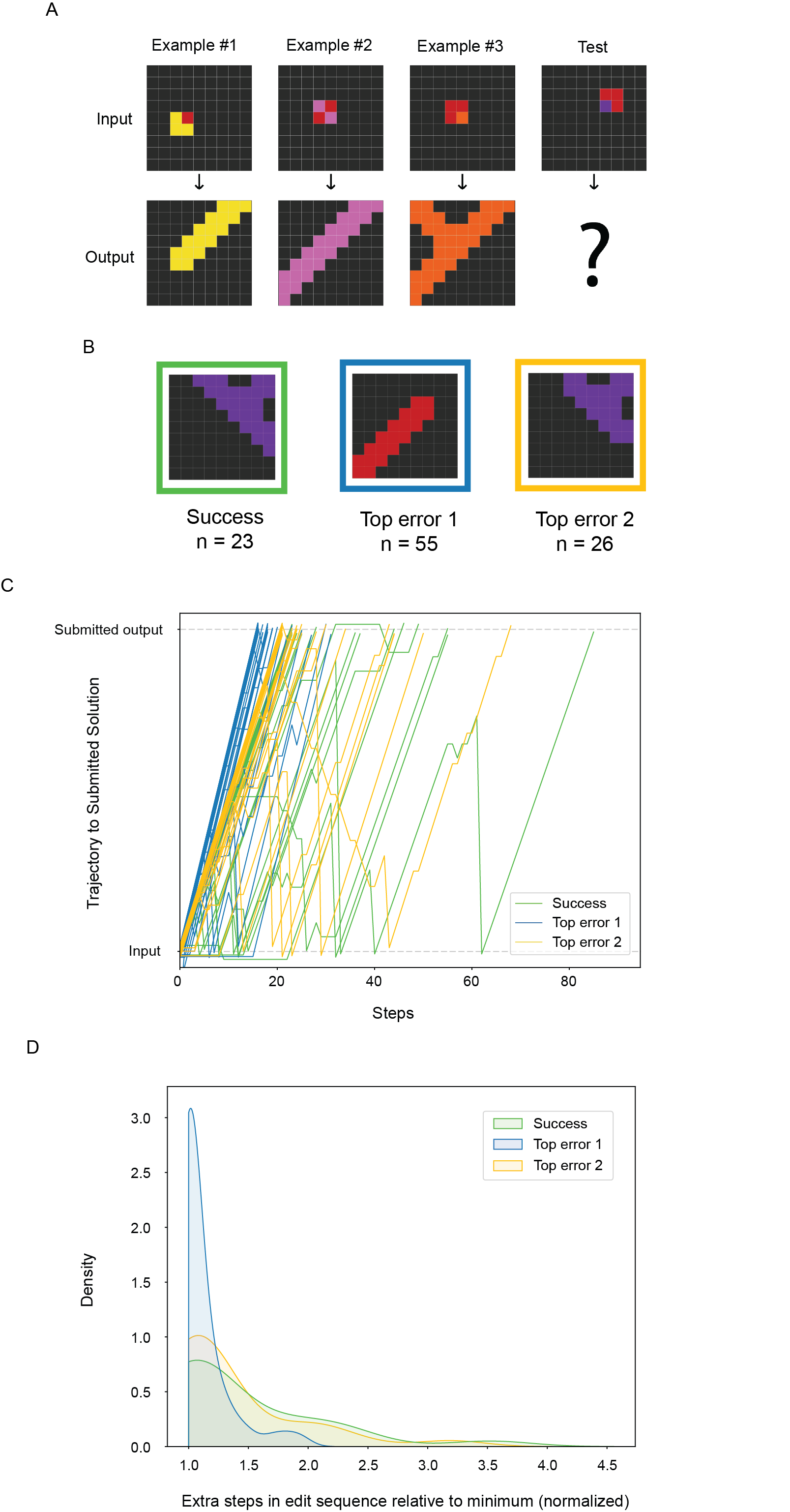}
  \end{minipage}\hfill
  \begin{minipage}[c]{0.31\textwidth}
    \caption{
       Efficient success with greater variability among common errors in a path-extension problem. A) Input–output examples and test input. The correct transformation required replacing the red cells with purple paths extending diagonally. B) Most common first-attempt outputs. Each output is shown with a colored frame indicating solution type. Each of the two top incorrect outputs was produced by more participants than the correct solution. C) Edit-distance trajectories from input to submitted output. Each line represents one participant’s edit sequence, plotted as normalized distance from their submitted solution at each step, allowing comparison across solution types on a shared y-axis. A fixed vertical jitter was applied so that overlapping trajectories could be visualized. Participants who reached the correct solution on their first attempt typically followed short, direct trajectories, whereas common error trajectories were typically longer and more variable. D) Distribution of normalized extra steps relative to the minimum possible edit length. Successful solutions are tightly clustered near the minimum, while incorrect solutions show a wider spread in sequence length, indicating greater variability in the paths leading to common errors.
    } \label{fig:s3}
  \end{minipage}
\end{figure}

\end{document}